\documentclass[letterpaper, 10 pt, conference]{ieeeconf}  

\IEEEoverridecommandlockouts                              

\usepackage{epsfig} 
\usepackage{mathptmx} 
\usepackage{times} 
\usepackage{amsmath} 
\usepackage{amssymb}  
\usepackage{color}
\usepackage{xcolor}
\usepackage{preambles}
\usepackage{amsmath}
\usepackage{graphicx}
\usepackage{url} \usepackage{empheq}
\usepackage[format=plain,labelsep=period]{caption}
\usepackage{wrapfig}
\usepackage[list=on,listformat=simple]{subcaption}
\usepackage{hyperref}
\usepackage{url}
\usepackage{caption}
\usepackage{subcaption}
\hypersetup{colorlinks = true, citecolor=black}
\usepackage{todonotes}
\usepackage{alip_variables}
\usepackage[switch]{lineno} 

\title{\LARGE \bf
Moving past point-contacts: Extending the ALIP model to humanoids with non-trivial feet using hierarchical, full-body momentum control
}
\author{Victor C. Paredes$^{1}$, Daniel A. Hagen$^{2}$, Samuel W. Chesebrough$^{2}$, Riley Swann$^{2}$, \\Denis Garagi\'c$^{2}$, Ayonga Hereid$^{1}$
\thanks{*This work was supported by a Sarcos Technology and Robotics Corp. Research Grant.}%
\thanks{$^{1}$Mechanical and Aerospace Engineering, The Ohio State University, Columbus, OH, USA. {\tt\footnotesize (paredescauna.1, hereid.1)@osu.edu.}}
\thanks{
$^{2}$Sarcos Technology and Robotics Corp., Salt Lake City, UT, USA. {\tt\footnotesize d.hagen@sarcos.com.}
}%
}

\usepackage[capitalise]{cleveref}

\usepackage[noadjust]{cite}

\begin{document}

\maketitle
\thispagestyle{empty}
\pagestyle{empty}

\begin{abstract}
The Angular-Momentum Linear Inverted Pendulum (ALIP) model is a promising motion planner for bipedal robots. 
However, it relies on two assumptions: (1) the robot has point-contact feet or passive ankles, and (2) the angular momentum around the center of mass, known as centroidal angular momentum, is negligible. 
This paper addresses the question of whether the ALIP paradigm can be applied to more general bipedal systems with complex foot geometry (e.g., flat feet) and nontrivial torso/limb inertia and mass distribution (e.g., non-centralized arms).
In such systems, the dynamics introduce non-negligible centroidal momentum and contact wrenches at the feet, rendering the assumptions of the ALIP model invalid.
This paper presents the ALIP planner for general bipedal robots with non-point-contact feet through the use of a task-space whole-body controller that regulates centroidal momentum, thereby ensuring that the robot's behavior aligns with the desired template dynamics.
To demonstrate the effectiveness of our proposed approach, we conduct simulations using the Sarcos\textsuperscript{\copyright}~Guardian\textsuperscript{\textregistered} XO\textsuperscript{\textregistered} robot, which is a hybrid humanoid/exoskeleton with large, offset feet. 
The results demonstrate the practicality and effectiveness of our approach in achieving stable and versatile bipedal locomotion.
\end{abstract}

\section{Introduction}
\label{sec:intro}

\subsection{Motivation}
Humanoid robots have long been studied and are of particular interest for their ability to navigate complex and challenging terrains as humans would.
However, bipedal locomotion is not easily achievable as these systems are under-actuated\footnote{More degrees of freedom than there are actuators.} and therefore prone to falling as they repeatedly come in and out of contact with the ground. 

Attempts to model and plan stable gait trajectories using these complex, nonlinear humanoid-robot dynamics usually result in limited walking modes that lack proper feedback control, making them fragile to disturbances and un-modeled dynamics. 
Alternatively, a simplified template model for gait can be used to allow for more scalable gaits to be computed in real time and can include feedback when certain conditions are met.
However, these conditions do not hold true for general bipedal robots, limiting the use of these feedback-based gait planners.
This paper modifies one such template (the Angular Momentum-based Linear Inverted Pendulum or ALIP model) and removes its limiting conditions to allow for it to be generalized to all bipedal robots.

\subsection{Related Work}
Linear Inverted Pendulum (LIP) models of bipedal gait are often used as templates to simplify a robot's dynamics and formulate center of mass (CoM) trajectories that keep the center of pressure (CoP) of the system within the support foot polygon\cite{kajita20013d}.
These methods are useful for generating gaits in real-time but lack proper feedback control, which leads to limited-speed, conservative gaits. 

Alternatively, by considering passive ankles and a fixed step duration, the resulting under-actuated system exhibits naturally linear behavior such that its states at the end of each step can be predicted.
This allows for these LIP-based walking planners to incorporate feedback based on real-time predictions and foot placements to produce periodically stable gait cycles~\cite{xiong20223, gong2022zero, paredes2020dynamic}.
However, this passive ankle condition is more readily met for robots with (near) point-contact feet, which makes the use of these planners difficult to implement on general robots with complicated foot geometry (e.g., flat feet).
Additionally, humanoids with non-trivial torso/limb inertia (e.g., those with non-centralized arms and heavier limbs) need explicit momentum regulation to allow for the centroidal dynamics to be approximated by these simplified LIP models.
For this paper, we choose to expand upon the Angular Momentum-based Linear Inverted Pendulum (ALIP) model/planner since it is reported to present numerical advantages over other models like the Hybrid Linear Inverted Pendulum (H-LIP)~\cite{gong2022zero,xiong20223}.
We will show that by using an appropriate task-space control formulation (that regulates the centroidal momentum and the center of mass position) the ALIP model can be applied to more general bipedal robots, thus removing the limiting assumptions needed to use this model. 

Task-space controllers for bipedal robots must handle the floating base coordinates in order to deal with the full-order robot dynamics while enforcing contact constraints.
Moreover, the control objectives must be defined such that the system can handle them while respecting any physical restrictions.
Leveraging these tasks and constraints can be realized by formulating a quadratic-programming based controller such as \cite{nakanishi2008operational, reher2020inverse, wensing2022optimization}, in which the tasks are appropriately weighted so as to not interfere with the constraints (which does not guarantee the completion of certain tasks or constraints).
Another option is to use hierarchical quadratic programming whereby tasks/constraints are solved sequentially in the null space of any previous tasks such that any subsequent objectives cannot violate the optimality of preceding objectives~\cite{herzog2014balancing,herzog2016momentum, del2014prioritized,del2014partial}.  
This paper extends upon the control structure presented in \cite{del2014partial} as it produces a fast computation time while explicitly handling prioritized tasks. 

\subsection{Contribution}
This paper presents the following contributions:
\begin{itemize}
    \item We analyze and provide the conditions needed for \emph{any} bipedal robot to have its full dynamics captured by the ALIP model.
    \item We formulate a task-space hierarchical controller that enforces these conditions as tasks which enables the full-order robot dynamics to be approximated by the simplified ALIP dynamics.
    \item We then extend the ALIP planner (which is typically applied to point-contact robots with lumped torso mass and lighter limbs) to robots with flat feet, non-trivial torso/limb inertia, and larger non-centralized arms.
\end{itemize}

\noindent The effectiveness of our proposed approach is demonstrated in simulation, where the Sarcos\textsuperscript{\copyright}~Guardian\textsuperscript{\textregistered} XO\textsuperscript{\textregistered} robot (a hybrid humanoid/exoskeleton with large, offset feet and non-centralized arms) is made to follow a variety of stable gait patterns.





\begin{figure}[b]
\vspace{-1em}
\centering
\includegraphics[width=\linewidth]{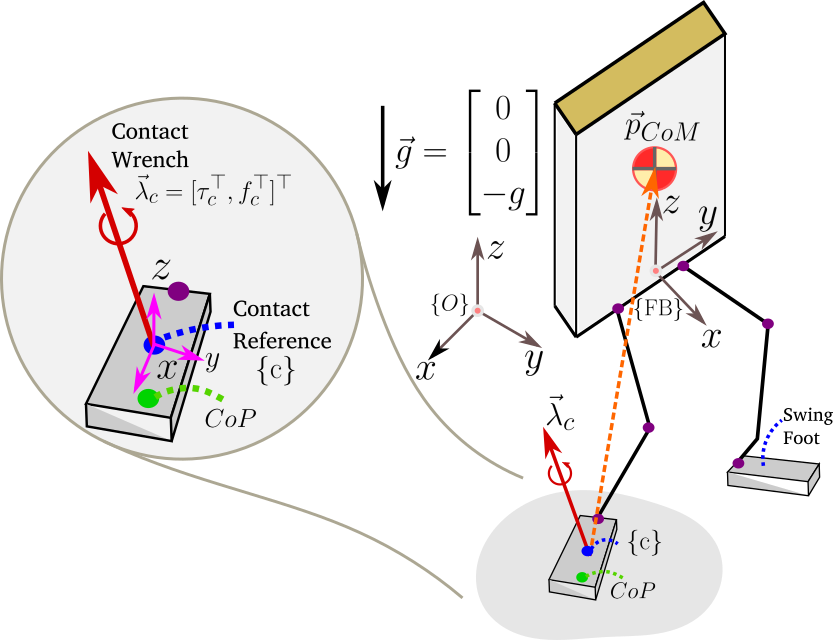}
\caption{During the single support phase, the bipedal robot will experience a resolved contact wrench $\lamC=[\lammC^\top~\lamfC^{\hspace{0.2em}\top}]^\top$ at the reference contact frame $\{c\}$ (assuming that the \emph{no-slip} contact constraint is maintained). 
}
\label{fig: RobotFrames}
\end{figure}

\section{ALIP model}
\label{sec:alip_model}

\subsection{Floating Base Humanoid Dynamics}
\label{sec:robot_dynamics}

Consider the following rigid body dynamics for a floating base (FB) humanoid system:

\begin{equation}\label{eq:fb_dynamics}
    M\ddqvec + C(\qvec,\dqvec) + G(\qvec) - J_c^\top \lamC = S^\top\vec{\tau}
\end{equation}

\noindent where $M$, $C$, and $G$ represent the mass matrix, Coriolis vector, and gravity vector, respectively, $\qvec$ is the configuration of the system (FB position/orientation + joint angles), $\vec{\tau}$ is the torque applied to each degree-of-freedom (with $S$ being a selector matrix to denote that the FB is not actuated), $J_c$ is the contact Jacobian and $\lamC$ are the generalized contact wrench(es). 
In the general case, the robot makes contact with the ground using one or both feet at the contact point(s) $c$, such that the contact Jacobian can be written as:%
\begin{equation}\label{eq:contactJ}
    J_c \triangleq S_c \left[J_{c,LF}^\top~J_{c,RF}^\top\right]^\top
\end{equation}%
\noindent where $S_c$ selects the active contacts.
This imposes a holonomic constraint of the form:
\begin{align}\label{eq:stick-slip}
    J_c \ddqvec + \dot{J}_c \dqvec = 0
\end{align} 
\noindent which ensures the current position(s) and orientation(s) of the contact(s) are held constant (i.e., the \emph{no-slip} condition).

\subsection{Centroidal Dynamics}
The ALIP model is commonly applied to bipedal robots with point-contact feet, where (in single support) the contact point is assumed to (i) coincide with the CoP and (ii) have zero reaction moments~\cite{vukobratovic2004zero,dunn1996foot,pratt1996virtual,kajita1991study}. 
However, for more general robots with non-point-contact feet, the CoP location can change during the gait cycle and is not easily measured or estimated. 
We instead project the centroidal dynamics to a user-defined, arbitrary contact frame $\{c\}$ (consistent with the contact Jacobian, $J_c$) located within the foot polygon.
Moreover, we orient the contact frame $\{c\}$ \emph{for each step} such that the $x-$ and $y-$axes coincide with the forward and lateral directions of motion for the center of mass, respectively, as illustrated in Fig. \ref{fig: RobotFrames}.
Once the calculations are carried out in this arbitrary frame, they are re-expressed in global coordinates to be used in the control formulation.

During single support, the dynamics of a bipedal robot can be represented in $\{c\}$ by the angular momentum at the contact point (or \emph{contact angular momentum}; $\LcontC$) and the relative vector to the center of mass (CoM; $\pcomC$)\footnote{We use $\vec{x}^{\hspace{0.1em}a}$ to denote a vector in the reference frame $a$. When no superscript is used, the World frame is assumed (unless otherwise noted).}.
Using the parallel axis theorem, we relate the contact angular momentum to the \emph{centroidal angular momentum} (or angular momentum of the CoM; $\LcomC$) via: %
\begin{align}
    \LcontC &= \LcomC + \pcomC \times (m \vcomC) \label{eq:angmom_CoP1} \\
    &= \LcomC + m \begin{bmatrix} p_y \dot{p}_z - p_z \dot{p}_y \\ p_z \dot{p}_x - p_x \dot{p}_z \\ p_x \dot{p}_y - p_y \dot{p}_x \end{bmatrix} \label{eq:angmom_CoP2}
\end{align}
\noindent where $\pcomC = [p_x~p_y~p_z]^\top$ and $\vcomC = [\dot{p}_x~\dot{p}_y~\dot{p}_z]^\top$ denote the relative position and velocity of the CoM w.r.t. the contact point, respectively, and $m$ denotes the mass of the system. 
As the remaining derivations are carried out in the contact frame, we will ignore the superscript notation for simplicity (e.g., $\LcontC := \vec{L}_{c} = [L_{cx}~L_{cy}~L_{cz}]^\top$). 

\begin{figure}[t]
\centering
\includegraphics[width=\linewidth]{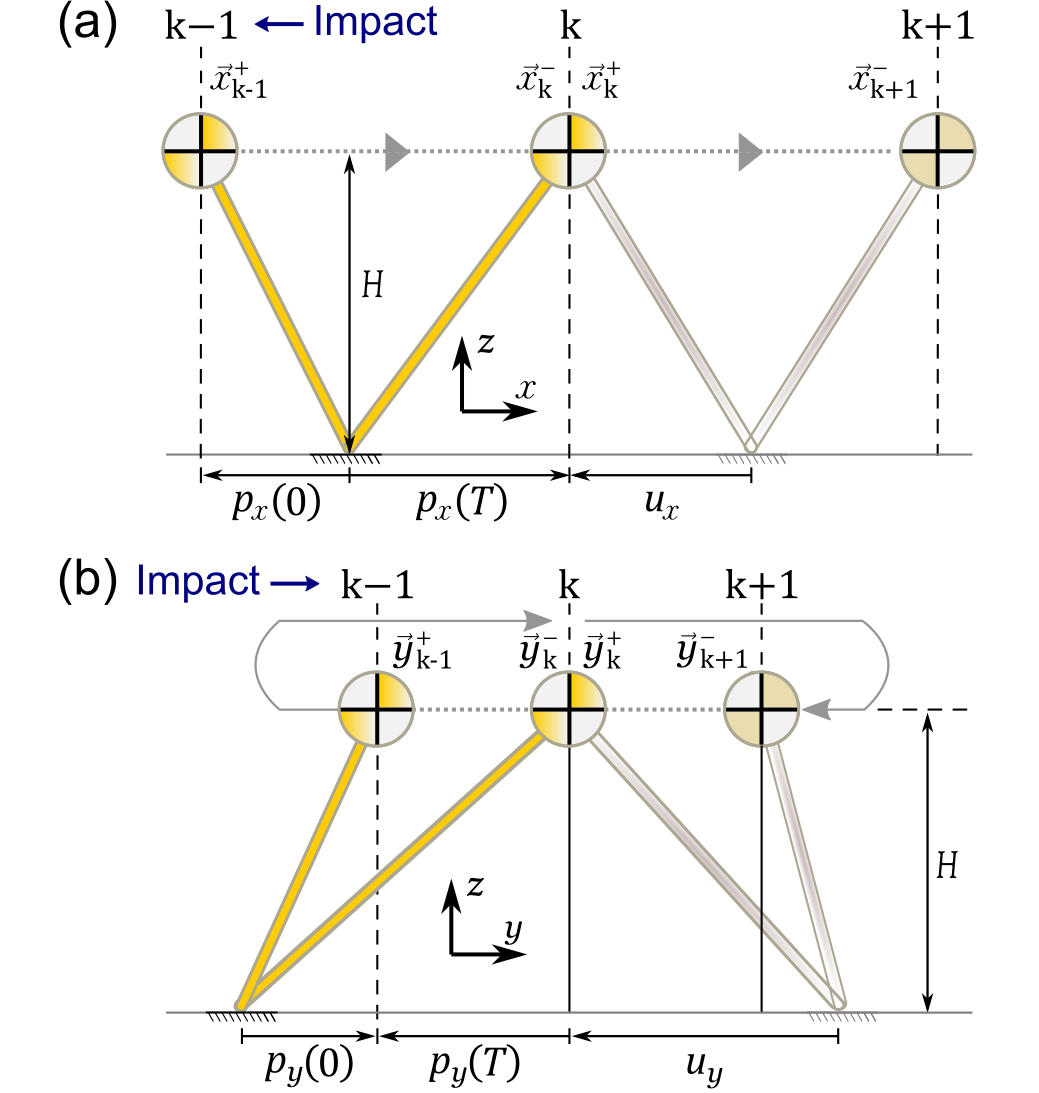}
\caption{Representation of the ALIP model walking with positive velocity in the sagittal and frontal planes for two consecutive steps delimited by the impacts $\{k-1, k, k+1\}$ with the states before (-) and after the impact (+). Ideally the states converge such that $\vec{x}_k^{\hspace{0.2em}-} = \vec{x}_{k+1}^{\hspace{0.2em}-}$ and $\vec{y}_k^{\hspace{0.2em}-} = \vec{y}_{k+1}^{\hspace{0.2em}-}$.}  
\label{fig: steps}
\vspace{-1em}
\end{figure}

\subsection{ALIP Dynamics and Planner}
In general, two main assumptions are made in order to formulate the ALIP dynamics and planner.
A first and reasonable assumption is that the CoM height is held constant (i.e, $p_z = H, \dot{p}_z = 0$). 
This allows us to modify (\ref{eq:angmom_CoP2}) to obtain the horizontal equations of motion by solving for the velocities $\dot{p}_x$ and $\dot{p}_y$,
\begin{empheq}[left=\empheqlbrace\hspace{0.5em}]{align}
\dot{p}_x&=\frac{L_{cy}}{m H} - \frac{L_{CoMy}}{m H} \label{px_ALIP_nl} \\[0.5em]
\dot{p}_y&=-\frac{L_{cx}}{m H} + \frac{L_{CoMx}}{m H} \label{py_ALIP_nl}
\end{empheq}
\noindent where, $L_{CoMi}$ is the centroidal angular momentum about the $i^{\text{th}}$ axis.



The second assumption is that the centroidal angular momentum $\Lcom$ is negligible compared to the contact angular momentum $\Lcont$.
This is a reasonable assumption for point-contact robots with legs that can be considered mass-less and torso/arm inertia that are not substantial, but for more complicated bipedal systems this is not always true. 
In Section \ref{sec:controller}, we discuss how the task space momentum controller can enforce these conditions, but for now we continue on with the derivations assuming these conditions have been met. 

As a consequence of choosing a contact frame that is not generally coincident with the CoP, the contact frame will experience non-zero reaction moments during the gait cycle.
As such, the rate of contact angular momentum for a given step is given by:
\begin{equation}\label{eq:LcRate}
    \LcontDot = \vec{p}_{CoM} \times m \grav + \lammC
\end{equation}
\noindent where $\lammC$ are the (potentially non-zero) reaction moments at $\{c\}$.
For robots with point contacts (where the contact frame is located near the CoP) it is reasonably assumed that the reaction moments are zero, which reduces (\ref{eq:LcRate}) into a linear system that only depends on the position of the CoM. 
Traditionally, these assumptions allow for (\ref{px_ALIP_nl})--(\ref{eq:LcRate}) to be rewritten as linear systems that represent the reduced order ALIP model which can then be used to plan for future states of the robot during gait.

For the purpose of planning the future states of more complicated robots without point-contacts, we choose to ignore these contact moments as well.
In reality, the true rate of contact momentum (as well as the rate of centroidal angular momentum) for these more complicated systems will depend on these non-zero reaction moments. 
In Section \ref{sec:controller}, we define additional control tasks such that (i) the robot is made to follow the future states as defined by the reduced order ALIP model and (ii) that the reaction wrench is limited by physical constraints to enforce contact conditions (e.g., \emph{no-slip}). 
In this way, by utilizing the full reaction wrench to minimize $\Lcom$ and to track the future states of the robot while adhering to physical constraints, the system is made to behave similarly to the reduced order ALIP model even with non-zero contact moments.

By defining the states $\xvec = [p_x~L_{cy}]^\top$ and $\yvec = [p_y~L_{cx}]^\top$  the ALIP dynamics can be represented as motion in the frontal and sagittal planes, respectively.%
\begin{align}
    \dot{\vec{x}} &= \begin{pmatrix}
        \dot{p}_x \\
        \dot{L}_{cy}
    \end{pmatrix} = \begin{bmatrix}
        0 & \frac{1}{mH}\hspace{0.3em} \\
        \hspace{0.7em}mg & 0 
    \end{bmatrix}\begin{pmatrix}
        p_x \\
        L_{cy}
    \end{pmatrix}=A\xvec\label{eq:ALIP-x} \\
    \dot{\vec{y}} &= \begin{pmatrix}
        \dot{p}_y \\
        \dot{L}_{cx}
    \end{pmatrix} = \begin{bmatrix}
        0 & \frac{-1}{mH}\hspace{0.3em} \\
        -mg & 0 
    \end{bmatrix}\begin{pmatrix}
        p_y \\
        L_{cx}
    \end{pmatrix}=-A\yvec\label{eq:ALIP-y}
\end{align}%
Solutions to (\ref{eq:ALIP-x}) \& (\ref{eq:ALIP-y}) can be defined as $\xvec(t) = M_x(t) \xvec(0)$ and $\yvec(t) = M_y(t) \yvec(0)$, respectively, where:
\begin{align}
    M_x(t) &= \begin{bmatrix} \cosh(\ell t) & \frac{1}{mH\ell} \sinh(\ell t) \\ mH \ell \sinh(\ell t) & \cosh(\ell t)\end{bmatrix}\\[0.5em]
    M_y(t) &=\begin{bmatrix} \cosh(\ell t) & -\frac{1}{mH\ell} \sinh(\ell t) \\ -mH \ell \sinh(\ell t) & \cosh(\ell t)\end{bmatrix}
\end{align}
\noindent and $\ell = \sqrt{g/H}$.
These equations explicitly define the horizontal motion of the CoM and the contact angular momentum (about the horizontal axes) for the ALIP model at time $t$, given some initial state of the system at the beginning of each step (Fig. \ref{fig: steps}).

\begin{figure}[t!]
\centering
\includegraphics[width=\linewidth]{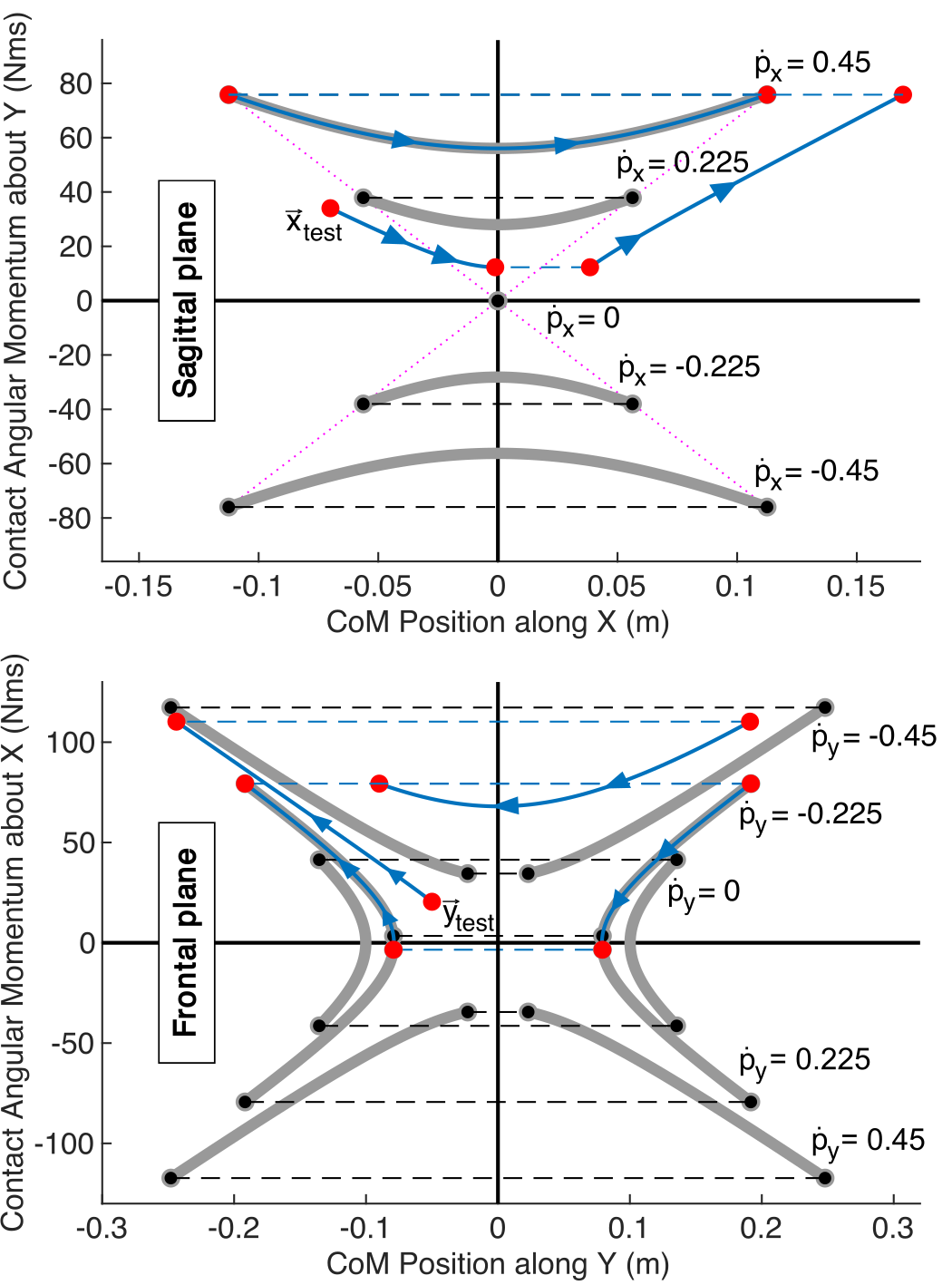}
\caption{Phase portrait for the sagittal and frontal plane of motion. The ALIP template produces stable hybrid orbits represented by the thick grey lines for the continuous part of the dynamics and the dashed lines for the instantaneous impact. Note that the contact angular momentum is impact invariant. The blue lines show the effect of using the feedback regulation $(u_x, u_y)$ to drive an initial state (here $x_{test}$ and $y_{test}$) to a desired orbit.}
\label{fig: ALIP_PP}
\vspace{-1em}
\end{figure}

As the motion of the ALIP dynamics is completely defined by the initial states of the system, the ALIP planner can then fundamentally rely on the estimation of the contact angular momentum at the end of the \emph{current} step $(\hat{L}_{cy}, \hat{L}_{cx})$ in order to calculate the desired landing position of the swing foot ($u_x, u_y$) such that the contact angular momentum at the end of the \emph{next} step can be regulated.
Tracking these desired foot-placements then allows the robot to track ALIP states that are periodically stable (Fig. \ref{fig: ALIP_PP}).
The state planning for each horizontal direction is calculated as follows:


\subsubsection{Sagittal plane}
We rely on the ALIP trajectories to relate the current state $\xvec(t) = [p_x(t)~L_{cy}(t)]^\top$ to the estimated contact angular momentum about the $y$-axis at the end of each step (after $T$ seconds).
\begin{align}
    \hat{L}_{cy} = mH \ell \sinh(\ell (T-t)) p_x(t) + \cosh(\ell (T-t)) L_{cy}(t)
\end{align}

For a desired forward velocity, $\dot{p}^{\hspace{0.1em}d}_x$, the desired contact angular momentum $L_{cy}^d$ is derived from the expected displacement between two impacts ($\dot{p}^{\hspace{0.1em}d}_x T = p_x(T) - p_x(0)$) which is then used to calculate the desired forward foot placement ($u_x$; Fig. \ref{fig: steps}\emph{a}).
\begin{empheq}{gather}
    L_{cy}^d= \frac{mH \ell \dot{p}^{\hspace{0.1em}d}_x T}{2} \left( \frac{1 + \cosh(\ell T)}{\sinh(\ell T)} \right)\\[0.7ex]
    u_x=\frac{L^d_{cy}-\cosh(\ell T) \hat{L}_{cy}}{mH \ell \sinh(\ell T)}
\end{empheq}

\subsubsection{Frontal plane}
Analogously, the estimated contact angular momentum about the $x$-axis at the end of each lateral step is a function of the current state $\yvec(t) = [p_y(t)~L_{cx}(t)]^\top$. 
\begin{align}
\hspace{-2mm}
    \hat{L}_{cx} = -mH \ell \sinh(\ell (T-t)) p_y(t) + \cosh(\ell (T-t)) L_{cx}(t)
\end{align}

For a desired lateral velocity, $\dot{p}^{\hspace{0.1em}d}_y$, the desired contact angular momentum $L_{cx}^d$ is derived from the expected displacement ($\dot{p}^{\hspace{0.1em}d}_y T=p_y(T)-p_y(0)$; Fig. \ref{fig: steps}\emph{b}) as:
\begin{align}
    L_{cx}^d &= -mH \ell \frac{\sinh(\ell T)}{1 + \cosh(\ell T)} p^* - mH \ell \coth(\ell T) \dot{p}^{\hspace{0.1em}d}_y T \\
    p^* &= 
    \begin{cases}
      \frac{W}{2} - \min(0, \dot{p}^{\hspace{0.1em}d}_y)T & \text{(Left Support)}\\
      -\frac{W}{2} - \max(0, \dot{p}^{\hspace{0.1em}d}_y)T & \text{(Right Support)}
    \end{cases}  \label{eq: py0}
\end{align}
\noindent where $W$ is the desired width between feet during in-place walking. Note that in \eqref{eq: py0} we make sure that the distance between the contact foot and the CoM at the beginning and end of each step is no less than $\frac{W}{2}$. 
This results in the lateral foot placement ($u_y$) given by:

\begin{align}
    u_y=- \frac{L^d_{cx}-\cosh(\ell T) \hat{L}_{cx}}{mH \ell \sinh(\ell T)}
\end{align}

The feedback actions $u_x$ and $u_y$ render the ALIP dynamics periodically stable in each plane of motion~\cite{gong2020angular}. 


\section{Task-space Hierarchical Momentum Control}
\label{sec:controller}

As previously mentioned, the ALIP planner assumes (i) a constant CoM height and (ii) negligible centroidal angular momentum compared to the contact angular momentum. 
While these assumptions may be more easily made for smaller bipedal systems with point-contact feet, it is difficult to do so for larger bipedal systems with complicated foot geometry, non-trivial torso/limb inertia, and larger moving arms.
In order to apply the ALIP template model and planner to these systems, a task-space controller must be used in order to explicitly control the centroidal momentum of the system while ensuring that foot contact constraints are not violated. 
For this, we specifically utilized a task-space hierarchical quadratic programming controller modified from \cite{del2014partial,herzog2014balancing}. 
In this way, the dynamics more closely resemble the ALIP model of a mass rotating about a contact point and the ALIP planner can be used to produce stable gait.



The task space controller primarily relies on the condition that the FB of the system is always controllable (i.e., $\text{rank}(J_c)\geq6$), which is satisfied when one or both feet are (i) in contact with the ground and (ii) motionless (\ref{eq:stick-slip}).
Inverting this no-slip condition produces an equation for joint accelerations that ensures solutions lie in the nullspace of this constraint,

\begin{equation}\label{eq:invStickSlip}
    \ddqvec = J_c^\dagger(-\dot{J}_c\dqvec) + Z_c\vec{z}_c
\end{equation}

\noindent where $Z_c = \text{Null}(J_c)$ and $\vec{z}_c$ is the optimization variable. 
Using a hierarchical quadratic program and (\ref{eq:invStickSlip}) it is then possible to solve for the value of $\vec{z}_c$ that satisfies any number of acceleration tasks/constraints while ensuring that any/all subsequent tasks do not violate the optimality of any preceding tasks (including the no-slip condition)~\cite{herzog2014balancing,herzog2016momentum, del2014prioritized,del2014partial}.

In order to formulate the centroidal momentum as an acceleration task, we derive the centroidal momentum matrix ($A_{CoM}$) and its derivative from \cite{wensing2015efficient} and then differentiate the centroidal momentum equation $\vec{h}_{CoM} = A_{CoM}\dqvec$, thus producing the equation for the rate of centroidal momentum.

\begin{equation}
    \dot{\vec{h}}_{CoM} = A_{CoM}\ddqvec + \dot{A}_{CoM}\dqvec
\end{equation}

\begin{figure}[t!]
\centering
\includegraphics[width=0.8\linewidth]{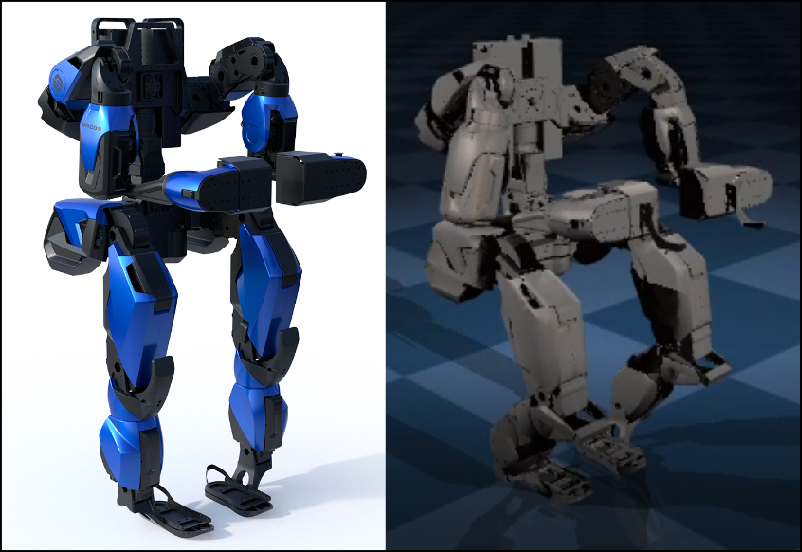}
\caption{Sarcos\textsuperscript{\copyright}~Guardian\textsuperscript{\textregistered} XO\textsuperscript{\textregistered} is a 150 kg full-body hybrid humanoid/exoskeleton robot with non-point-contact feet. We observe a rendered version on the left and the right, the MuJoCo model used in the simulation. 
}
\vspace{-4mm}
\label{fig: GuardianXO}
\end{figure}

\noindent Note, that $\vec{h}_{CoM}$ consists of $\vec{L}_{CoM}$ as well as the linear momentum of the CoM ($\vec{K}_{CoM}$), which can also be represented as the linear velocity of the CoM scaled by the mass of the robot.
Therefore, the desired rate of centroidal momentum can be formulated to control (i) for the desired horizontal CoM trajectories provided by the ALIP planner while (ii) minimizing centroidal angular momentum and (iii) keeping the CoM height constant. 
The desired rate of centroidal momentum is therefore defined as:

\begin{equation}\label{eq:desiredHgDot}
    \begin{aligned}
        \dot{\vec{h}}^d_{CoM} &= \begin{pmatrix}
            \textbf{0} \\
            m \ddot{\vec{p}}_{CoM,d}
        \end{pmatrix}\\[1ex]
        &\hspace{0.5em}+ K_D\left(
        \begin{pmatrix}
            \textbf{0} \\
            m \dot{\vec{p}}_{CoM,d}
        \end{pmatrix}
        - A_{CoM}\dot{\vec{q}}
        \right)\\[1ex]
        &\hspace{1.5em}+ K_P\left(
        \begin{array}{c}
            \textbf{0}\\
            m \vec{p}_{CoM,d} - m \vec{p}_{CoM,m}
        \end{array}
        \right)
    \end{aligned}
\end{equation}
\noindent where $\vec{p}_{CoM,m}$ represents the measured CoM position, $\vec{p}_{CoM,d}$ and its derivatives represent the desired trajectories generated by the ALIP model (\ref{eq:ALIP-x}) and (\ref{eq:ALIP-y}) (with constant CoM height), and $K_D$ and $K_P$ and are appropriate positive definite gains.

During single support, if we consider the contact frame $\{c\}$ to be the only contact (along with the no-slip condition), the wrench at that point is fully determined from the desired rate of centroidal momentum.
This desired rate may not be physically achievable by the system (e.g., the contact wrench may break \emph{stick-slip} conditions).
We therefore calculate the \emph{constrained} desired rate of centroidal momentum by constraining this wrench to adhere to physical bounds (i.e., friction and torque limits based on foot geometry) which limits the CoP to stay within the contact polygon.
By closing on this constrained rate of centroidal momentum, the controller can track the desired ALIP trajectories without violating contact conditions.

Lastly, we define an additional task to control for the position/orientation of the swing foot (which is updated by the ALIP parameters $u_x$ and $u_y$) as well as a task to maintain some ideal configuration of the arms.

\section{Results}
\label{sec:results}


\subsection{Robot description}
Sarcos\textsuperscript{\copyright}~Guardian\textsuperscript{\textregistered} XO\textsuperscript{\textregistered} is a full-body industrial hybrid exoskeleton/humanoid as shown in Fig. \ref{fig: GuardianXO}. 
This 150 kg robot has feet that are offset from the ankle joint centers that make it impossible to consider these as point-contacts or to apply any passive ankle schemes to recover behavior similar to point-contact feet.
Additionally, the legs cannot be considered mass-less and the arms and torso substantially contribute to the momenta of the system, making this a perfect candidate to test our integrated task-space momentum controller with ALIP planner. 



\renewcommand{\arraystretch}{1.5}
\begin{table}[t!]
\centering
\begin{tabular}{||c | c | c | c | c | c||} 
 \hline
 Time (s) & 0-2 & 2-8 & 8-14 & 14-20 & 20-22 \\  
 \hline\hline
  $\dot{p}_x^{des}$ (m/s)& 0 & 0.225 & 0.45 & 0.225 & 0 \\[0.5ex]
  \hline
 $\dot{p}_y^{des}$ (m/s) & 0 & 0 & 0 & 0 & 0  \\ [0.5ex] 
 \hline \noalign{\vspace{2ex}}
 \hline
 Time (s) & 22-28 & 28-30 & 30-36 & 36-38 & 38-44 \\
 \hline\hline
 $\dot{p}_x^{des}$ (m/s)& 0 & 0 & -0.225 & 0 & 0  \\[0.5ex]
 \hline
 $\dot{p}_y^{des}$ (m/s) & -0.225 & 0 & 0 & 0 & 0.225\\[0.5ex]
 \hline
\end{tabular}
\caption{Desired steps in horizontal CoM velocities.}
\label{tab:desiredVel}
\vspace{-1em}
\end{table}
\renewcommand{\arraystretch}{1}

\subsection{Forward and lateral walking}
To demonstrate the effectiveness of the task-space hierarchical momentum controller, simulations were conducted in MuJoCo \cite{todorov2012mujoco} whereby the robot was made to autonomously walk in either the sagittal or frontal planes of motion at variable speeds using the ALIP planner. 
In particular, the robot was made to follow the step changes in commanded horizontal velocity provided in Table \ref{tab:desiredVel}.

\begin{figure}[b]
\centering
\includegraphics[width=87.13611mm]{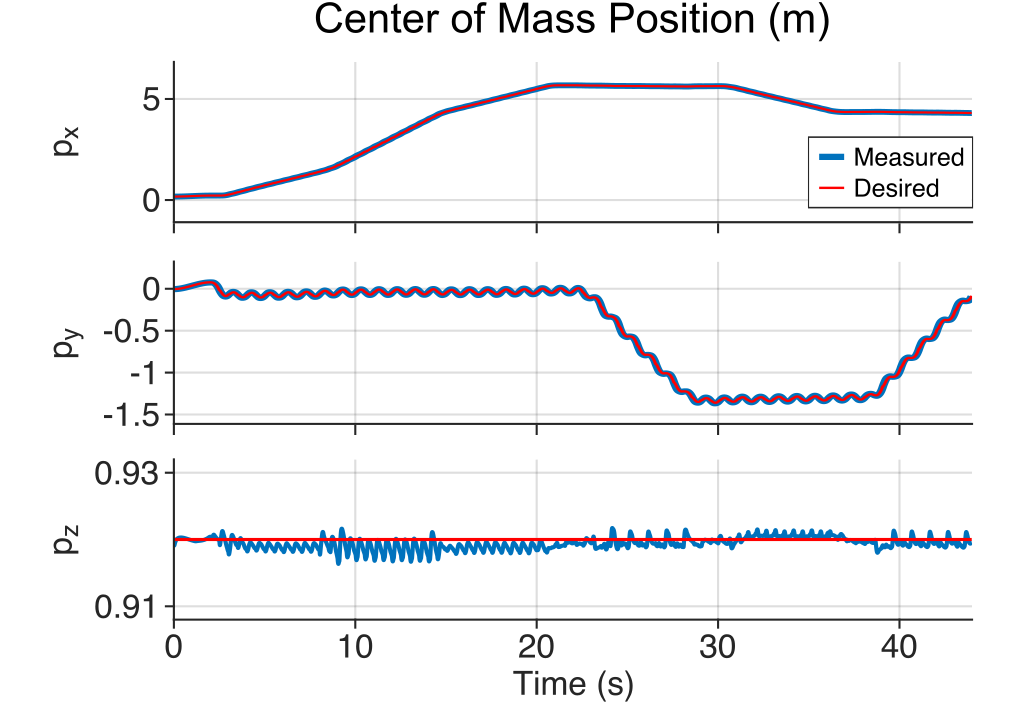}
\caption{
Comparing the desired CoM trajectories (\emph{red}) prescribed by the ALIP planner with the measured CoM values produced by the centroidal momentum controller (\emph{blue}). 
}
\label{fig:CoMPosition}
\end{figure}

\begin{figure}[b]
\centering
\includegraphics[width=87.13611mm]{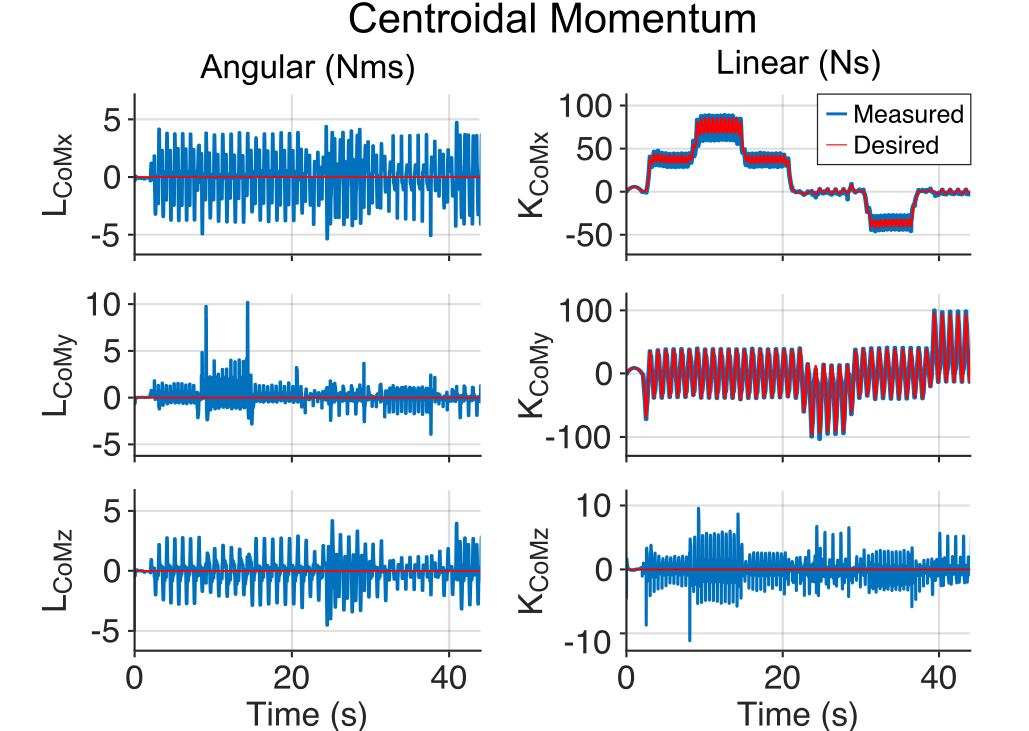}
\caption{
Desired (\emph{red}) vs. measured (\emph{blue}) centroidal momentum for various steps in desired CoM velocity.
}
\vspace{-3mm}
\label{fig:momentumTracking}
\end{figure}

\begin{figure}[t]
\centering
\includegraphics[width=87.13611mm]{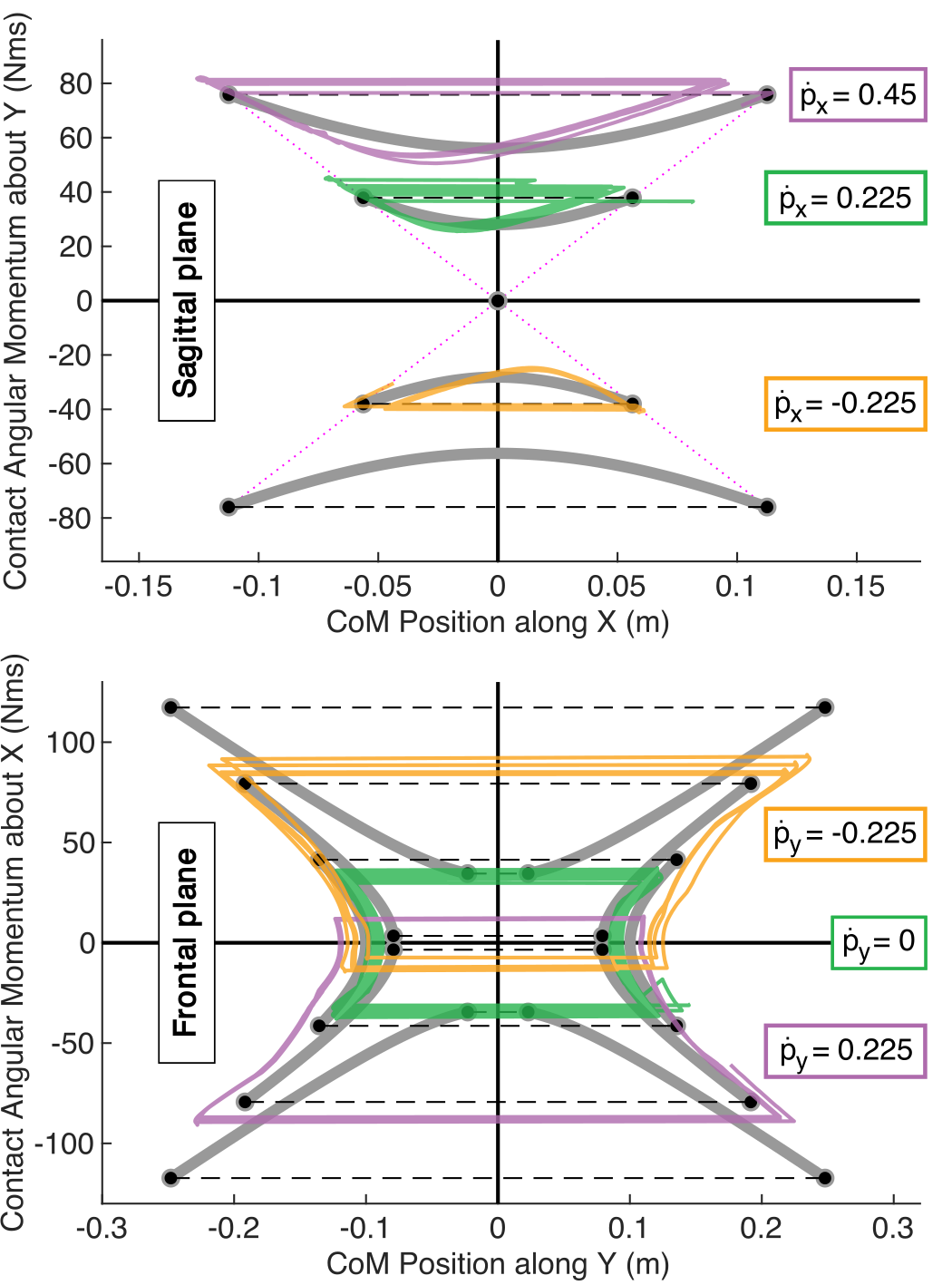}
\caption{Measured ALIP states overlaid on the ideal ALIP phase portraits in the sagittal and frontal planes as the robot tracks various horizontal speeds (See Table \ref{tab:desiredVel}).
The measured states as the robot transitions between target velocities have been omitted.} 
\vspace{-1em}
\label{fig:PP}
\end{figure}

\begin{figure}[t]
\centering
\includegraphics[width=87.13611mm]{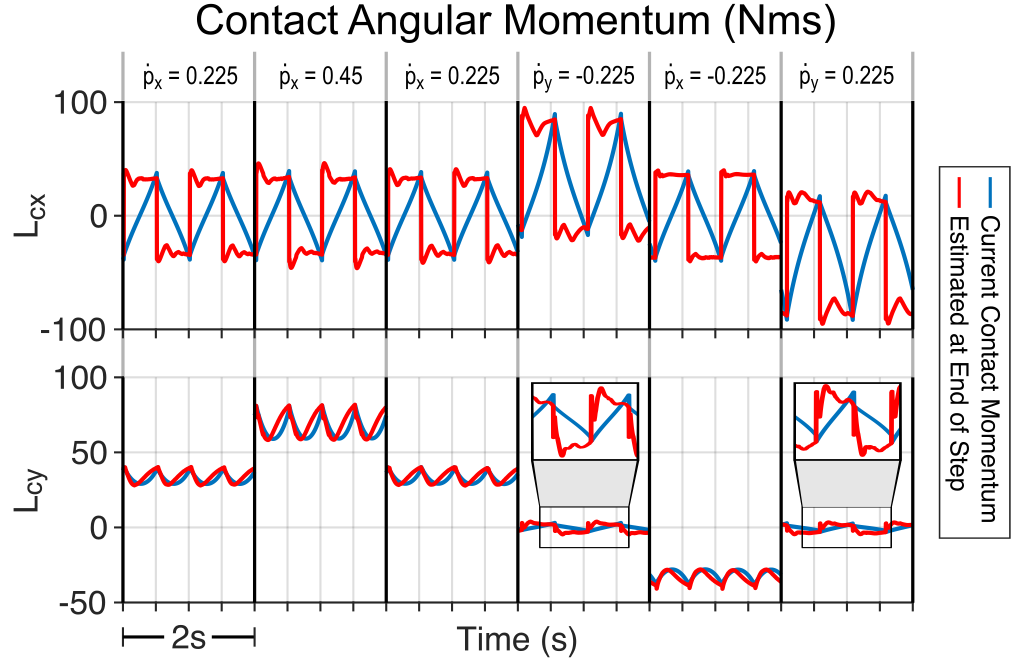}

\caption{Current contact angular momentum (\emph{blue}) vs. the predicted values at the end of each step (\emph{red}). For each 2 second window of the non-zero desired CoM velocities from Table \ref{tab:desiredVel}, we find that the expected and measured values converge at the end of each step. 
After the robot makes contact with the ground it exhibits a brief peak in angular momentum, attributed to a discrete jump in vertical CoM velocity, causing the expected and measured values at the start of each step to deviate. 
 }
 \vspace{-1em}
\label{fig:contactMomentumEst}
\end{figure}



To render the full-order dynamics closer to the ALIP model, our hierarchical task-space controller must (i) minimize the centroidal angular momentum while (ii) maintaining a constant CoM height and (iii) following the desired horizontal ALIP CoM trajectories. 
Our contact frame $\{c\}$ lies in the center of the contact foot and is aligned such that $x$ is forward and $z$ is up.
We see in Fig. \ref{fig:CoMPosition} that the controller reliably tracks the prescribed lateral CoM position while maintaining a constant height.
Alternatively, Fig. \ref{fig:momentumTracking} demonstrates that the controller is able to sufficiently track the horizontal linear momentum (i.e., the CoM velocity) while minimizing (i) the centroidal angular momentum and (ii) the linear momentum of the CoM along the $z$-axis (to maintain a constant CoM height). 

Once the robot reaches a desired CoM velocity, the measured ALIP states behave in a manner consistent with the ALIP template, which can be seen when those values are overlaid on the respective phase portraits from in Fig. \ref{fig:PP}. 
This demonstrates that by minimizing the centroidal angular momentum and tracking the CoM trajectories as prescribed by the ALIP planner, the robot dynamics can indeed be captured by the ALIP template model (even though the reaction moments are non-zero).
Additionally, we find that the contact momentum estimates at the end of each step are close to the measured values (an important assumption for calculating the feedback actions for periodic gait; Fig. \ref{fig:contactMomentumEst}), which further strengthens the claim that the full-robot model is being captured by the ALIP dynamics when controlled via this momentum controller.
Note that following contact with the ground, the robot exhibits a brief peak in angular momentum, attributed to a discrete jump in vertical CoM velocity, which causes a discrepancy between actual and estimated contact momentum at the start of each step. 
Observe that the resulting centroidal angular momentum in Fig. \ref{fig:momentumTracking} is about an order of magnitude less than the resulting contact angular momentum in Fig. \ref{fig:contactMomentumEst}, consistent with the assumptions of the ALIP model. 

\begin{figure}[h]
\centering
\includegraphics[width=87.13611mm]{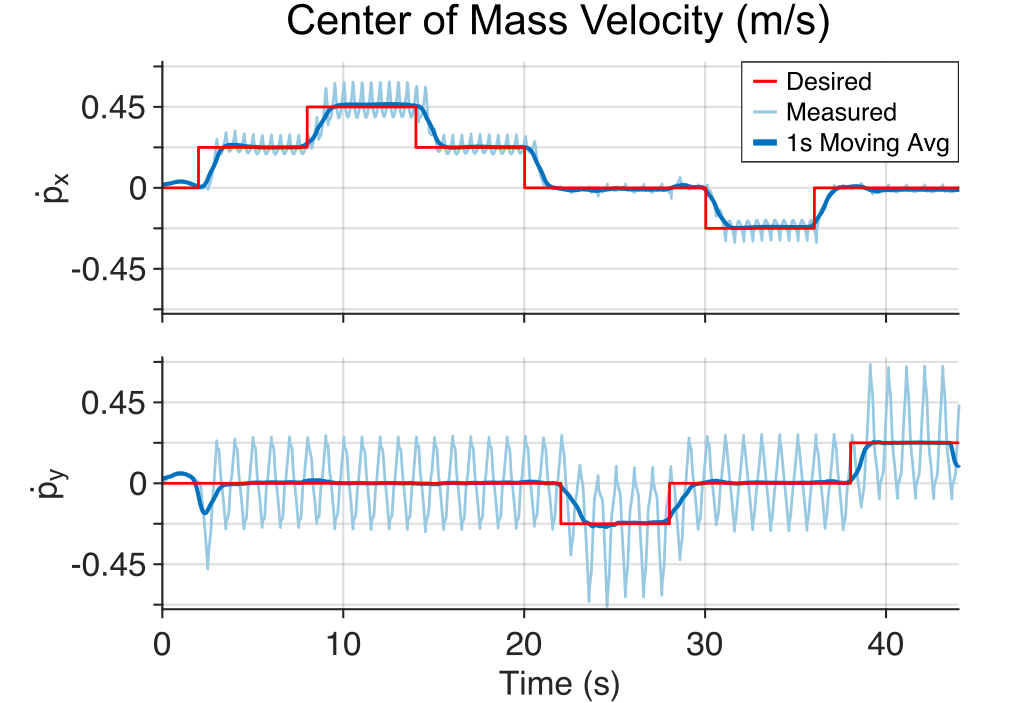}
\caption{
Commanding various desired CoM speeds using the ALIP planner with a centroidal momentum controller. 
Note that for either forward/backward or side-to-side movements, the controller is able to adequately track the desired linear velocity of the CoM. 
}
\vspace{-1em}
\label{fig:VelocityTracking}
\end{figure}

Lastly, as the feedback actions rely on some desired horizontal velocity, it is important to explore how well the controller is able to track these desired velocities.
Using the values from Table \ref{tab:desiredVel}, we compare the desired and measured horizontal CoM velocities in Fig. \ref{fig:VelocityTracking}. 
Note that in order to achieve a large speed command, smaller intermediate speeds are used to avoid large swing foot movements which can be mathematically correct but physically too demanding for the robot.
When comparing the desired velocities to a 1 second moving average of the measured values, we find that the controller does a remarkable job at tracking the desired velocity in both the forward and lateral directions.

\section{Conclusions}
\label{sec:conclusion}
The main highlight of this paper is the extension of the ALIP paradigm to robots with non-point-contact feet, non-trivial torso/limb inertia, and non-centralized arms. 
We have shown that the full-order dynamics of the Sarcos\textsuperscript{\copyright}~Guardian\textsuperscript{\textregistered} XO\textsuperscript{\textregistered} can be captured by the ALIP dynamics when a task-space controller is formulated to minimize the centroidal angular momentum and to track the CoM references prescribed by the ALIP planner while regulating contact wrenches/constraints.
We demonstrate that this new formulation is able to accurately predict robot states based on the ALIP paradigm even though this robot does not have point-contact feet and, as a result, non-zero reaction moments at the defined contact frame.

Under the conditions imposed by the controller, we were then able to apply the ALIP planner, which, produced stable longitudinal and lateral gaits of variable speeds.
Specifically, we were able to obtain a maximum of 0.45 m/s in the forward direction (slightly more than 1 mph) and 0.225 m/s in either lateral directions. 
We therefore show that the fundamental assumptions previously needed to use ALIP models (i.e., point contact feet, massless limbs, and lumped torso mass) can be relaxed when the planner is combined with an appropriately-constrained centroidal momentum controller.
These results, therefore, expand on the applicability of the ALIP template model to a more general type of bipedal robot when combined with an appropriate task space controller.

\IEEEtriggercmd{\enlargethispage{-5.15in}}

\bibliographystyle{IEEEtran}
\bibliography{references.bib}

\end{document}